The use of large language models to enhance cancer clinical trial educational materials

Mingye Gao, MS [1*]; Aman Varshney, MS [2*]; Shan Chen, MS [3,4]; Vikram Goddla [3,4], Jack Gallifant, MBBS[3,4]; Patrick Doyle, MS[4]; Claire Novack, BS[4]; Maeve Dillon-Martin, BS[4]; Teresia Perkins, BS[4]; Xinrong Correia[5]; Erik Duhaime PhD[5]; Howard Isenstein, MA[6], Elad Sharon, MD[7]; Lisa Soleymani Lehmann, MD PhD[8]; David Kozono, MD PhD[4]; Brian Anthony, PhD [1]; Dmitriy Dligach, PhD[9]; Danielle S. Bitterman, MD [3,4*]

1. Massachusetts Institute of Technology, Cambridge, MA, USA
2. Technical University of Munich, Munich, Germany
3. Artificial Intelligence in Medicine (AIM) Program, Mass General Brigham, Harvard Medical School, Boston, MA, USA
4. Department of Radiation Oncology, Brigham and Women's Hospital/Dana-Farber Cancer Institute, Boston, MA, USA
5. Centaur Labs, Boston, MA, USA
6. Digidence, Bethesda, MD, USA
7. Department of Medical Oncology, Dana-Farber Cancer Institute, Boston, MA, USA
8. Department of Medicine, Mass General Brigham, Harvard Medical School, , Boston, MA, USA
9. Loyola University Chicago, Chicago, IL, USA

*Co-first authors

Corresponding Author:
Danielle Bitterman, MD
Dr. Danielle S. Bitterman
Department of Radiation Oncology
Dana-Farber Cancer Institute/Brigham and Women's Hospital
75 Francis Street, Boston, MA 02115
Email: dbitterman@bwh.harvard.edu
Phone: (857) 215-1489
Fax: (617) 975-0985

**Links to Supplemental Materials:**
Appendix Tables and Figures: 📄 Appendix

**Conflicts of interest:**
DK: Advisory and consulting, unrelated to this work: Genentech/Roche.
DSB: Editorial, unrelated to this work: Associate Editor of Radiation Oncology, HemOnc.org (no financial compensation); Advisory and consulting, unrelated to this work: MercurialAI.
HI: Principal, unrelated to this work; Digidence.
LSL: Employee of Verily;  Advisor unrelated to this work: MediSensor Technologies, BellSant
XC: Employee of Centaur Labs
ED: Employee of Centaur Labs

**Acknowledgements:**
The funder did not play a role in the design of the study; the collection, analysis, and interpretation of the data; the writing of the manuscript; and the decision to submit the manuscript for publication.
The authors would like to acknowledge the BROADBAND Research Project at the Brigham & Women's Hospital Department of Radiation Oncology for providing regulatory and personnel support for this project. The BROADBAND Project was in part made possible by the generous donations of Stewart Clifford, Fredric Levin, and their families.

**Funding:**
The authors acknowledge financial support from the Google PhD Fellowship (SC), the Woods Foundation (DB, SC, JG), the NIH (NIH R01CA294033 (SC, JG, ES, DK, DB), NIH U54CA274516-01A1 (SC, DB), NIH R01LM012973 (DD)), and the American Cancer Society and American Society for Radiation Oncology, ASTRO-CSDG-24-1244514-01-CTPS Grant DOI #: https://doi.org/10.53354/ACS.ASTRO-CSDG-24-1244514-01-CTPS.pc.gr.222210 (DB). This work was also conducted with support from UM1TR004408 award through Harvard Catalyst | The Harvard Clinical and Translational Science Center (National Center for Advancing Translational Sciences, National Institutes of Health) and financial contributions from Harvard University and its affiliated academic healthcare centers. The content is solely the responsibility of the authors and does not necessarily represent the official views of Harvard Catalyst, Harvard University, and its affiliated academic healthcare centers, or the National Institutes of Health. The authors thank Google Cloud research fund for Claude API inference costs.

**ABSTRACT**

Cancer clinical trials often face challenges in recruitment and engagement due to a lack of participant-facing informational and educational resources. This study investigated the potential of Large Language Models (LLMs), specifically GPT4, in generating patient-friendly educational content from clinical trial informed consent forms. Using data from ClinicalTrials.gov, we employed zero-shot learning for creating trial summaries and one-shot learning for developing multiple-choice questions, evaluating their effectiveness through patient surveys and crowdsourced annotation. Results showed that GPT4-generated summaries were both readable and comprehensive, and may improve patients' understanding and interest in clinical trials. The multiple-choice questions demonstrated high accuracy and agreement with crowdsourced annotators. For both resource types, hallucinations were identified that require ongoing human oversight. The findings demonstrate the potential of LLMs "out-of-the-box" to support the generation of clinical trial education materials with minimal trial-specific engineering, but implementation with a human-in-the-loop is still needed to avoid misinformation risks.

**INTRODUCTION**

Clinical trials are the gold standard for investigating management strategies that can potentially improve cancer patient outcomes. The experimental nature of clinical trials necessitates clear information and effective communication about potential benefits and risks that patients could realize during the study period. However, patients have expressed needing better resources to learn about their trial options [1-4]. Currently, the primary resource by which cancer patients and providers learn about clinical trial options are clinical trial registries, such as ClinicalTrials.gov[5-6]. ClinicalTrials.gov is a large, public database of clinical trials, but it primarily uses highly technical language that is geared toward a clinician and investigator audience, meaning they are often inaccessible to most patients. Therefore, there is still a lack of widely accessible materials to inform and educate potential participants about specific clinical trial options. This challenge is an important barrier to trial recruitment, informed consent, protocol adherence, and successful and timely accrual. At a more fundamental level, ensuring adequate information about clinical trials is imperative to ensure valid informed consent and widening access to treatment options within a clinical trial across sociodemographic backgrounds.

Large Language Models (LLMs) present a new opportunity to enhance trial processes via improved patient awareness and engagement. Prior studies have investigated LLMs to facilitate cancer patient education and communication, demonstrating promise but also risks arising from falsifications and fabrications [7-11]. However, most work on using LLMs to improve clinical trial processes has focused on clinical trial matching [3-4, 12-13] and little research has investigated these models for enhancing informational and educational resources. LLMs' ability to simplify and summarize texts, in particular, is an exciting avenue for improving education and awareness. For example, LLMs could be used to simplify and clarify the often complex and jargon-heavy information presented in clinical trial documents, including informed consent forms [14-15]. In addition, LLMs could support the development of new methods to support clinical trial education. For example, established methods to measure the quality of clinical trial understanding after informed consent, such as the validated Quality of Informed Consent questionnaire [16], do not assess trial-specific details. LLMs could facilitate the development of trial-specific questionnaires, providing opportunities for patients to self-assess their understanding and provide a pathway for patient-specific education to address knowledge gaps.

In this study, we explored the potential and risks of using LLMs to generate informational and educational resources about clinical trials via the secondary use of clinical trial materials. First, we prompted LLMs to generate short, plain-language summaries of the key elements of a clinical trial from informed consent forms (ICFs). Patient surveys of these summaries investigated their readability and their potential to enhance clinical trial understanding as a part of informed consent. Second, we investigated the ability of LLMs to automatically generate multiple-choice question-answer pairs (MCQAs) based on ICF content, providing an interactive approach to assess clinical trial understanding. Our findings provide proof-of-concept for the potential of leveraging LLMs to enhance informational and education resources for cancer clinical trials, which could improve patient engagement and support clinical trial processes.

**METHODS**

Figure 1 illustrates our overall research approach of prompting GPT4 to generate new educational and informational resources for clinical trials: plain-language short summaries (to inform patients about a trial) and multiple-choice questions (to assess understanding of a trial).

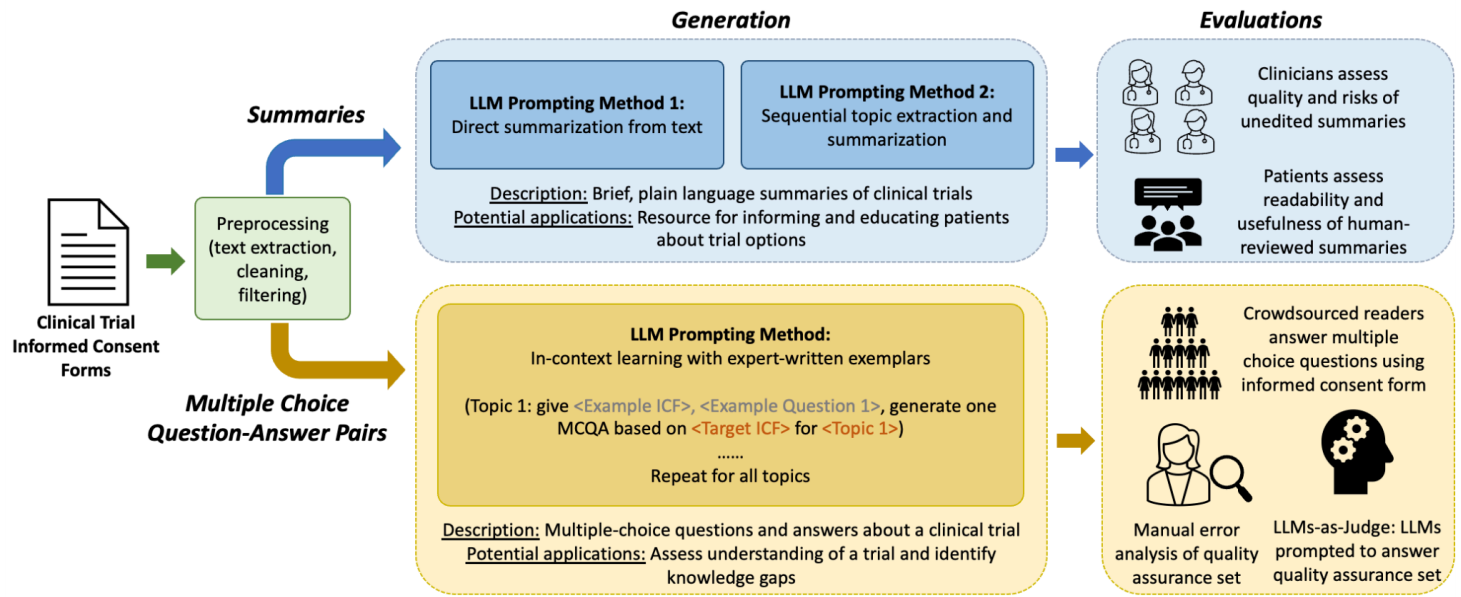

**Figure 1.** Illustration of the overall study design, including the approach to generating and evaluating summaries (blue) and multiple choice question-answer pairs (yellow) from clinical trial informed consent forms. LLM = large language model.

📄 TrialEducationFigure1.png

*Datasets*

ICFs for this study were collected using the ClinicalTrials.gov API [17]; PyMuPDF [18], a publicly available Python library, was used to retrieve text from the PDF files.

For the summary generation task, 11 clinical trial ICFs were randomly selected from ClinicalTrials.gov for prompt engineering and initial evaluation by the research team (Appendix A, Table 1). The statistics of this set are shown in Appendix A, Figure 1. A larger set of ICFs were used for the large-scale questionnaire development. We selected 91 interventional cancer clinical trials registered between January 1st, 2021 and April 15th, 2024. This time frame was chosen to capture the most up-to-date practices in informed consent while still providing a substantial pool of studies for analysis. The distribution of ICF pages and number of tokens are shown in Appendix A, Figure 2.

*Summary Generation*

We explored 2 different approaches to generating the trial summary from consent forms: Direct summarization from text, and sequential extraction and summarization. In the direct summarization approach, the raw ICF text was provided to GPT*4-0125-preview* for summary generation, with prompt instructions to include key information about the clinical trial, which were informed by the basic elements of informed consent in Subpart A of the Revised Common Rule [19] (Appendix A, Figure 3). In the sequential summarization approach, the LLM was first prompted to extract the sections of text describing basic elements of informed consent informed by Subpart A of the Revised Common Rule from the raw ICF text (Appendix A, Table 2 and Appendix A, Figure 4A). These extracted text sections were then used to generate a summary (Appendix A, Figure 4B) with temperature and top_p setting to 0.

*Summary Evaluation*

To evaluate the quality of our two prompting approaches, 4 clinicians from the research team (JG, LL, DK, DB) evaluated the 11 clinical trial summaries generated using the 2 aforementioned prompting methods. Each summary was evaluated for the binary presence or absence of key trial elements, readability, inaccuracies, biases, and hallucinations. Overall summary quality was assessed using a 5-point Likert scale. The full survey is in Appendix B.

Next, to explore patient perspectives of the generated summaries in a real-world setting, we invited patients undergoing informed consent for the BROADBAND Research Study (hereafter referred to as BROADBAND), a prospective secondary use protocol in the Department of Radiation Oncology at Brigham and Women's Hospital/Dana-Farber Cancer Institute, to participate in a survey evaluating 5 LLM-generated cancer clinical trial summaries generated using the sequential summarization approach, including a summary of BROADBAND plus 4 other trials representing an observational trial, Phase I trial, Phase I/II trial, and Phase III trial (Appendix A, Table 3). To avoid any risks of misinformation, all summaries underwent manual review and editing by an oncologist. In addition to minimizing any patient risk, this mimics the real-world application of such a summary, where a member of the clinical trial research team would review the GPT4-generated content before it reaches the patient. Participants completed the survey after the BROADBAND informed consent discussion, which provided the opportunity to assess the impact of the summary on understanding of a clinical trial for informed consent. The survey tool was developed in RedCap and participants could complete the survey on paper, a laptop in the clinic, or via an emailed link. The survey instrument is in Appendix C. Informed consent for the survey was waived as this study was deemed to be exempt human subjects research by the Mass General Brigham Institutional Review Board (MGB Protocol # 2024P000949).

*Multiple choice question-answer pair generation*

GPT-4-1106-preview was used for MCQAs generation. To ensure the quality of the generated MCQAs, we adopted the in-context learning method: when generating an MCQA for each topic based on a target ICF, we fed an expert-created question-answer pair and its corresponding ICF text, along with the target ICF text, into GPT4. 15 MCQAs focused on a subset of the basic elements of informed consent were manually written by a board-certified oncologist (DB) based on an exemplar ICF for a non-cancer clinical trial [20] (Table 1). Using these pairs as in-context examples, we engineered a multi-turn prompt to generate MCQAs (Appendix A, Figure 5). For each generation query, temperature and top_p were set to 0, and max_tokens was set to 3000. After filtering invalid generations (e.g., responses returned by GPT4 that were not an MCQA), a total of 1335 MCQAs were generated for 91 ICFs.

| Topic | Questions | Multi-Choice Options | Answer |
|---|---|---|---|
| A statement that the study involves research | True or False: This study involved research | A) True; B) False | A |
| An explanation of the purposes of the research | What program is being evaluated in this research study? | A) A new blood pressure medication; B) The STROKE Telemedicine Outpatient Program (STOP); C) Referral to a clinic neurologist only; D) Referral to a primary care physician only | B |
| The expected duration of the subject's participation | How long will each participant be involved in this study? | A) 6 months; B) 3 months; C) 12 months; D) 36 months | A |
| A description of the procedures to be followed | If you are randomized to the STOP-Stroke group, what will occur at 1 week, 1 month, 3 months, and 5 months after enrollment? | A) Your blood pressure will be checked; B) You will be asked to complete a questionnaire; C) You will have a video visit with a nurse practitioner or stroke physician, social worker, and pharmacist and will receive educational text messages every other week; D) You will have a visit with your neurologist and primary care provider | C |
| The approximate number of subjects involved in the study | About how many patients will be enrolled in the study? | A) 25; B) 50; C) 75; D) 100 | D |
| Identification of any procedures which are experimental | What is the experimental intervention in this study? | A) Follow up with primary care to monitor your blood pressure every 2 weeks; B) 24-hour blood pressure monitoring each month; C) The STOP program: video visits with a stroke prevention team and educational text messages; D) Educational text messages every other week alone | C |
| A description of any reasonably foreseeable risks or discomforts to the subject | The following is a possible risk of this study. | A) Breach of confidentiality; B) Worsening blood pressure; C) Increased risk of stroke; D) None of the above | A |
| A description of any benefits to the subject or to others which may reasonably be expected from the research | Select the benefits of participating in the study (you may select more than one). | A) You will receive educational materials about blood pressure control after stroke; B) Your blood pressure will be better controlled; C) Your stroke risk will be lower; D) You will receive 24-hour blood pressure monitoring. | A, D |
| A disclosure of appropriate alternative procedures or courses of treatment, if any, that might be advantageous to the subject | What is the alternative option to participating in the study? | A) You will be offered a visit in the stroke clinic and will follow-up with your primary doctor; B) You will have regular video visits with a stroke specialist; C) There is no alternative to participating; D) You will receive regular education about reducing your stroke risk | A |
| A statement describing the extent, if any, to which confidentiality of records identifying the subject will be maintained | If you participate in this study, who may review your personal health information? You may select more than one | A) Representatives of the University of Texas Health Sciences Center at Houston; B) The research sponsor; C) No one; D) Researcher from other hospitals | A, B |

| | option. | | |
|---|---|---|---|
| For research involving more than minimal risk, an explanation as to whether any compensation, and an explanation as to whether any medical treatments are available, if injury occurs and, if so, what they consist of, or where further information may be obtained | What treatment will be available for you if you are injured as a result of participating in the study? | A) No treatment; B) All needed facilities, emergency treatment and professional services will be made freely available; C) All need facilities, emergency treatment and professional services will be made available, but not free of charge; D) None of the above | C |
| Research, Rights or Injury: An explanation of whom to contact for answers to pertinent questions about the research and research subjects' rights, and whom to contact in the event of a research-related injury to the subject | Who can you contact with any questions, concerns, or input about the study? | A) The principal investigator and members of the study team; B) No one; C) Administrators at the University of Texas Health Sciences Center at Houston; D) Your primary physician | A |
| A statement that participation is voluntary, refusal to participate will involve no penalty or loss of benefits to which the subject is otherwise entitled, and the subject may discontinue participation at any time without penalty or loss of benefits, to which the subject is otherwise entitled | If you enroll in the study, when can you choose to stop participating? | A) You cannot stop participating once you enroll; B) Only within 1 week of enrolling; C) At any time during the study; D) Before you complete the first questionnaire | C |
| Any additional costs to the subject that may result from participation in the research | Who will be charged for medications, studies, or procedures recommended to you during this study? | A) The study sponsor; B) You or your insurance, because they will be considered standard of care; C) UTHealth, because they are running the study; C) None of the above | B |
| The consequences of a subject's decision to withdraw from the research and procedures for orderly termination of participation by the subject | What will occur if you decide to stop being a part of the study? | A) You will not be able to continue receiving stroke care at UTHealth; B) There will be no change to the services available to you from UTHealth; C) You will not be eligible for clinical trials in the future; D) You will need to change your primary doctor | B |

**Table 1.** The oncologist-written Multi-Choice Question-Answers (MCQAs) that were used as in-context examples for automated MCQA generation with GPT4.

*Multiple choice question-answer pair evaluation*

Annotations of the GPT4-generated MCQAs were obtained using DiagnosUs, a medical annotation crowdsourcing platform (Centaur Labs, Boston, MA), which uses continuous performance monitoring and incentivization to optimize quality. Crowdsourced readers were provided the ICF for each set of MCQAs and instructed to answer the GPT4-generated question stems based on information present in the ICF. In total, 504 crowdsourced readers participated and included individuals with a range of clinical and non-clinical backgrounds; details on the breakdown of self-reported clinical training are reported in Appendix A, Table 4. The number of human readers answering each GPT4-generated MCQA questions stem is defined as qualified reads. Evaluations are reported using metrics defined in Table 2.

The large number of generated MCQAs precluded complete manual error analysis. Therefore, we defined a quality assurance set of 78 MCQAs with difficulty ≥ 0.6 and agreement ≤ 0.5 for manual error analysis (i.e., MCQAs with a majority of readers disagreement with the GPT4-assigned answer and with substantial disagreement between readers). Our goal with this quality assurance set was to gain a deeper understanding of why readers tended to disagree with the GPT4-assigned answer in order to understand failure modes. In addition, we used a multi-agent framework where one LLM verifies the GPT4-generated MCQAs for the quality assurance test set. GPT-4o, Cohere R+, Gemini Pro 1, and Claude 3 Sonnet were prompted to answer the MCQAs (Appendix A, Figure 6); temperature, top_p, and max_tokens were set to 0, 0 and 300, respectively, for the 4 LLMs.

| Term | Definition |
| --- | --- |
| **Qualified reads** | Number of crowdsourced readers answering each GPT4-generated MCQA question stem. |
| **Difficulty** | Qualified reads disagreeing with the GPT4 answer divided by total qualified reads. |
| **Agreement** | Qualified reads with the majority answer (i.e., the answer chosen by the majority of qualified reads) divided by total qualified reads. |
| **Accuracy** | Percentage of MCQAs where the majority answer matches the GPT4 answer. |

**Table 2.** Definitions of metrics and associated terms used in MCQA (multiple choice question-answer) evaluations.

**RESULTS**

*Clinician summary evaluation*

For the 11 summaries evaluated by clinicians, both prompting approaches achieved comparable results for readability and topic content (Figure 2). There was slightly less evidence of inaccuracies, biases, and hallucinations using the sequential prompting approach. Quality was rated as acceptable or better in the majority of responses using both prompting approaches (Figure 3), although results varied substantially across trials and evaluators. We found that inaccuracies and hallucinations tended to occur for topics that were not described in the given ICF. Summaries generated using the sequential prompting method were preferred in 38.6% (17/44) responses, and summaries generated using the direct prompting method were preferred in 61.4% (27/44) responses.

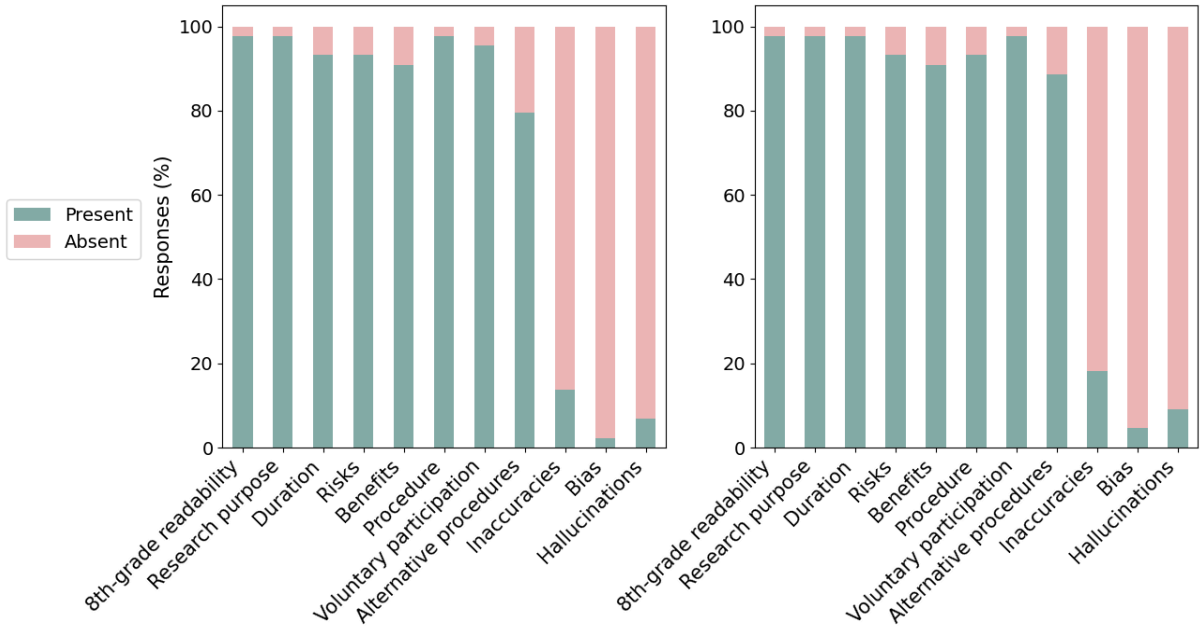

**Figure 2.** Clinician evaluation of 11 clinical trial summaries generated using the sequential prompting approach (left) and the direct prompting approach (right). There were 44 responses per item in the clinician evaluation summary.

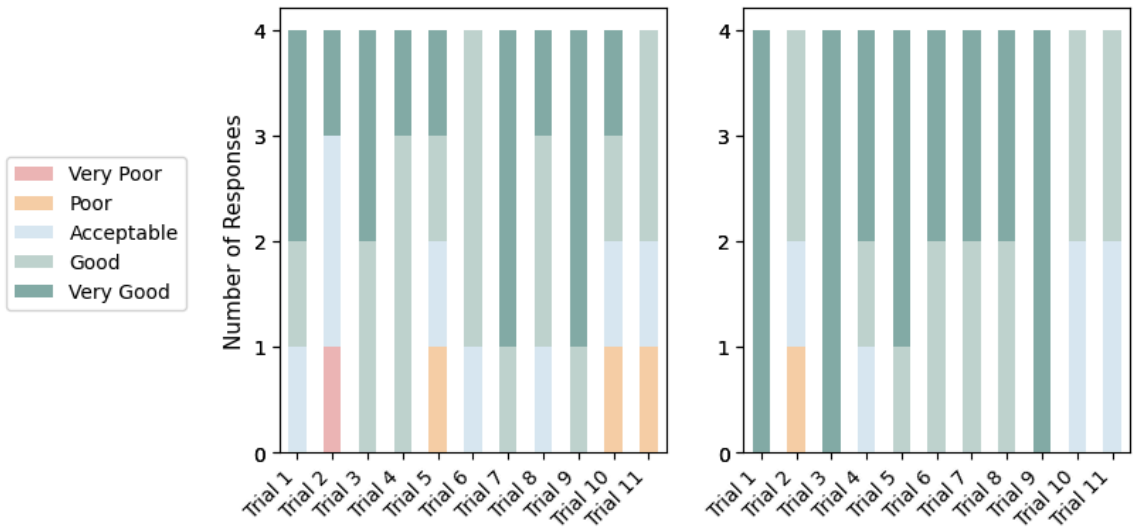

**Figure 3.** Clinician evaluation of overall quality of the 11 clinical trial summaries generated using the sequential prompting approach (left) and the direct prompting approach (right).

*Patient summary evaluation*

Thirteen patients completed the survey of the 5 cancer clinical trial summaries; Table 3 summarizes their characteristics, and Table 4 reports the key results. Complete results for each trial are in Appendix A, Table 5. Of note, a substantial portion of participants did not complete the survey for the non-BROADBAND trial summaries. Nevertheless, the majority of respondents found the summaries easy to understand and agreed or strongly agreed that the summaries provided sufficient information to decide about contacting the research team. Eleven of 13 (85%) of participants reported that the summary improved their understanding of the BROADBAND trial after having completed the informed consent process for the trial.

| Category | Subcategory | No. Survey Participants (N=13) |
|---|---|---|
| **Gender** | Male | 10 |
| | Female | 2 |
| | Missing | 1 |
| **Age (years)** | 31-40 | 1 |
| | 41-50 | 0 |
| | 51-60 | 3 |
| | 61-70 | 6 |
| | 71-80 | 3 |
| **Race** | White | 12 |
| | Asian | 1 |
| **Ethnicity** | Hispanic | 0 |
| | Non-Hispanic | 13 |
| **Prior clinical trial experience** | Yes | 4 |
| | No | 9 |

**Table 3.** Characteristics of clinical trial summary patient survey participants

| Survey Item | Strongly disagree | Disagree | Neither agree or disagree | Agree | Strongly agree | Missing |
|---|---|---|---|---|---|---|
| All Five Trial Summaries | | | | | | |
| This summary is easy to understand. | 0 | 3 (5%) | 3 (5%) | 18 (28%) | 26 (40%) | 16 (25%) |
| If I were researching trials, this summary provides enough information for me to decide if I would want to contact the research team to learn more about the trial. | 0 | 2 (3%) | 10 (15%) | 21 (32%) | 21 (32%) | 11 (17%) |
| BROADBAND Summary Only | | | | | | |
| I believe that reading this summary improved my understanding of BROADBAND | 0 | 0 | 2 (15%) | 6 (46%) | 5 (38%) | 1 (8%) |

**Table 4.** Key results of patient surveys evaluating 5 cancer clinical trial summaries, which were generated by GPT4 and manually reviewed and edited by an oncologist. Thirteen patients participated in the surveys. Complete results of the survey, presented for each trial individually, are in Appendix A, Table 5.

*Multiple choice question-answer pair evaluation*

MCQAs had an average of 5.21 qualified reads (standard deviation, 1.99). The majority answer agreed with the GPT4 answer in 1307/1335 (97.91%) of MCQAs, with an average agreement of 86.87% and an average difficulty of 15.52%. Of note, the median difficulty and agreement were 0.0 and 1.0, respectively, demonstrating that all readers' answers matched the GPT4-assigned answer in over half of all MCQAs. Appendix A, Table 6 and Appendix A, Figure 7 provide detailed statistics of qualified reads, agreement, and difficulty. When broken down by MCQA topic, average agreement was lowest and difficulty highest for MCQAs about contact information, expected duration, and alternative procedures (Appendix A, Figure 8).

We identified 78/1335 (5%) MCQAs with difficulty ≥ 0.6 and agreement ≤ 0.5 for our quality assurance set for manual error analysis. Manual review identified 5 error modes, which are summarized in Table 5. Of these, the majority of errors were human error, followed by errors in GPT4-generated MCQAs, errors due to missing information in the ICFs, and errors arising from ambiguous language.

For most MCQAs in the quality assurance set, the 4 LLMs disagreed with the MCQAs when there were incorrect GPT4-assigned labels, missing information in the ICF, and ambiguous definitions. This suggests that testing the generated MCQAs using other LLMs may be a promising avenue to assist humans in proofreading the quality of LLM-generated MCQAs. More details on the LLM and human reader results on the quality assurance set are shown in the Appendix A, Figures 9-12.

| Error Mode | Description | Number of MCQAs |
|---|---|---|
| Human Error | The MCQA is correct, and information needed to correctly answer it is present in the ICF. Of these, 24 MCQAs had an answer that was not difficult to identify in the ICF, and 3 MCQAs included a large number of details (e.g., >20 procedures, >1 page of potential risks) which made it particularly challenging to arrive at the correct answer. | 27 |
| Missing Information in ICF | The topic corresponding to the MCQA is not included or explicitly demonstrated in ICF. | 18 |
| Error in GPT4-generated MCQA | The GPT4-generated MCQA included more than one correct answer when only one correct answer was assigned (10) or the answer generated in the GPT4-g generated MCQA was incorrect (7). | 17 |
| Ambiguous Definition | The definition of term(s) in MCQA or ICF is not clear enough to point to the golden standard label (for example, ICF mentions "patients should inform their physician if they get injured during the study"; what does "physician" mean? Does it mean the principal investigator of the study or their primary care doctor?). | 13 |
| Not in English | ICF is not in English and readers cannot understand the MCQA | 3 |

**Table 5.** Error modes identified on qualitative error analysis of the quality assurance set.

**DISCUSSION**

This study explored the potential of LLMs to automatically draft patient-friendly summaries and MCQAs from ICFs. Experimental results demonstrated that LLMs can effectively draft readable and accurate summaries of ICFs with straightforward prompting techniques, with patient surveys suggesting potential roles in improving trial awareness and consent quality. Additionally, LLMs were able to generate high-quality MCQAs based on the content of target ICFs. These results provide proof-of-concept for leveraging LLMs to accelerate the development of new, diverse, and scalable educational resources for clinical trial patient education, while also highlighting key error modes that require ongoing human oversight and vigilance.

Prior studies have demonstrated the potential of leveraging LLMs for clinical education, including the generation of patient-facing disease-specific information [21-22], and summarization and simplification of existing clinical documents [23-33]. In similar work, White et al. (2023) used GPT3.5 to generate summaries of multiple clinical trials using the brief descriptions in ClinicalTrials.gov study records, although these were intended for researcher audiences [29]. In addition, studies have shown the ability of LLMs to simplify language in consent forms for standard medical procedures [34-37]. Our findings build on this prior literature to demonstrate two new, promising applications of LLMs to support the unique informational and educational needs of patients learning about clinical trials.

There is a need for better patient education about clinical trial options [38-40]. Inadequate educational resources about clinical trials limits awareness of and engagement in trials and may contribute to the high rate of cancer trials that fail to accrue [41-42]. New, diversified resources to educate patients about clinical trials could increase enrollment rates, improve patient understanding, and potentially broaden clinical trial access and diversity [43-49]. In fact, patients have expressed a need for more awareness about clinical research. During recruitment, patients and caregivers report a lack of familiarity with trial options and are more likely to have positive attitudes about participation if they learn about trials [3-4, 7]. Past studies have explored novel approaches to improve education, primarily at the informed consent stage [50-56], but scalability has previously been limited by the time and engineering expertise needed to develop trial-specific resources. Our findings show the potential of LLMs to lower the barrier to generating a diversity of educational resources from documents that are already created as a part of standard clinical trial conduct. Our prompting methods do not require significant engineering expertise to implement and are agnostic to the specifics of a given clinical trial.

While our findings demonstrate that GPT4 can, in general, follow prompt instructions to convert ICFs into new educational resources, they also highlight error modes necessitating ongoing human oversight and methods refinement. Though rare, both the summaries and MCQAs included inaccuracies and hallucinations, a known challenge of working with LLMs. These most often occurred when the ICF did not include adequate content requested in the prompt. This limitation may arise out of LLMs' alignment tuning, which leads them to prioritize helpfulness (i.e., following users' instructions in the prompt) over factual accuracy - a key error mode to monitor for such LLM applications. Our sequential prompting method for summaries, which first extracted relevant ICF text, and then summarized over the extracted text, appeared to mitigate but not completely alleviate this error mode. Further prompt refinement may improve these errors; however, human oversight is still needed to ensure accurate, comprehensive, and safe information when using LLMs for summarization [57]. At the same time, our findings may also spur trial sponsors to improve the content of ICFs.

Implementing such LLM applications for clinical trial processes, and in healthcare more broadly, is currently limited by a lack of effective means for large-scale evaluation and ongoing monitoring of model performance. As above, while promising, even state-of-the-art LLMs such as GPT4 require a human-in-the-loop to identify and resolve errors before they reach patients [58]. Nevertheless, our methods may lower the barrier to develop educational content because clinical trial staff may find it easier to review and revise LLM-generated drafts than to develop the content from scratch. While there is significant excitement about using LLMs as real-time conversational chatbots without a human-in-the-loop to support patient education, our study suggests that models "out-of-the-box" are not currently safe for such applications without oversight.

This study has several limitations that must be taken into account when considering our results. First, we evaluate a relatively limited number of trials and have a small number of human evaluators for the summaries, which limits generalizability. Additionally, the patients who agreed to participate in summary evaluations may not be reflective of the broader cancer population, including a lack of diversity. That said, ours is one of the very few studies that have assessed patients' perceptions of LLM outputs [59-60], and to our knowledge the only study evaluating patient perceptions of LLM content that relates to their own healthcare (i.e., their understanding of the BROADBAND study). Further, the number of human evaluators in our study falls within the range of other studies evaluating LLMs for patient education [61]. Nevertheless, given these limitations, our results should be considered as early proof-of-concept, and larger scale studies demonstrating safety, acceptability, and effectiveness are needed. In addition, we may not have used the optimal prompting approaches, and it is possible error rates could be reduced with additional prompt engineering. However, our goal was to understand the performance, behavior, and risks of widely available LLMs without significant additional engineering efforts, serving as a baseline for future technical innovation and advances. Similarly, including additional clinical trial materials, such as information from the ClinicalTrials.gov study records and trial protocols, may be a promising avenue to reduce the observed risk of inaccuracies and hallucinations by providing more substantive content on topics that may not be adequately described in an ICF. Finally, patients have diverse informational and educational needs. While we explored LLMs to generate two different types of educational resources, we did not explore personalizing the resources to individual preferences. This is an exciting avenue and will be a focus of future work.

**CONCLUSION**

Our findings demonstrate a promising role of LLMs in narrowing the knowledge gap between patients and specific clinical trial information, potentially enhancing informed decision-making and fostering greater patient engagement. This study establishes an initial framework for utilizing LLMs as supportive tools in patient-centered clinical trial education and informational resources. Future research should focus on optimizing output through advanced prompting techniques and automated oversight, investigating personalized approaches tailored to individual needs, and rigorously validating the quality, safety, and real-world impact and effectiveness of these methods.

# REFERENCES

[1] Kumar G, Chaudhary P, Quinn A, Su D. Barriers for cancer clinical trial enrollment: A qualitative study of the perspectives of healthcare providers. Contemp Clin Trials Commun. 2022 May 28;28:100939. doi: 10.1016/j.conctc.2022.100939. PMID: 35707483; PMCID: PMC9189774.

[2] BECOME Initiative Final Report. MBCA (Melanoma and Brain Cancer Alliance). Retrieved from https://www.mbcalliance.org/wp-content/uploads/BECOME-Final-Report-FULL.pdf

[3] Pretesting NIH clinical trial awareness messages: A focus study with patients, caregivers, and the general public. Bethesda, MD: National Institutes of Health, April, 2011.

[4] The need for awareness of clinical research [Internet]. National Institutes of Health (NIH). 2015 [cited 2024 May 30]. Available from: https://www.nih.gov/health-information/nih-clinical-research-trials-you/need-awareness-clinical-research

[5] N. L. of Medicine. ClinicalTrials.gov. https://clinicaltrials.gov/. [Accessed 23-11-2023]. 2023

[6] Nass S J, Moses H L, Mendelsohn J. Physician and patient participation in cancer clinical trials[M]//A National Cancer Clinical Trials System for the 21st Century: Reinvigorating the NCI Cooperative Group Program. National Academies Press (US), 2010.

[7] Chen S, Kann BH, Foote MB, et al. Use of Artificial Intelligence Chatbots for Cancer Treatment Information. *JAMA Oncol.* 2023;9(10):1459–1462. doi:10.1001/jamaoncol.2023.2954

[8] Chen S, Guevara M, Moningi S, et al. The effect of using a large language model to respond to patient messages[J]. The Lancet Digital Health, 2024, 6(6): e379-e381.

[9] Holstead R G. Utility of Large Language Models to Produce a Patient-Friendly Summary From Oncology Consultations[J]. JCO Oncology Practice, 2024: OP. 24.00057.

[10] Yeo Y H, Samaan J S, Ng W H, et al. Assessing the performance of ChatGPT in answering questions regarding cirrhosis and hepatocellular carcinoma[J]. Clinical and molecular hepatology, 2023, 29(3): 721.

[11] Zhu L, Mou W, Chen R. Can the ChatGPT and other large language models with internet-connected database solve the questions and concerns of patient with prostate cancer and help democratize medical knowledge?[J]. Journal of translational medicine, 2023, 21(1): 269.

[12] Joshi V, Kulkarni AA. Public awareness of clinical trials: A qualitative pilot study in Pune. Perspect Clin Res. 2012 Oct;3(4):125–132. PMCID: PMC3530979

[13] Jin Q, Wang Z, Floudas CS, Chen F, Gong C, Bracken-Clarke D, Xue E, Yang Y, Sun J, Lu Z. Matching Patients to Clinical Trials with Large Language Models. ArXiv [Preprint]. 2024 Apr 27:arXiv:2307.15051v4. PMID: 37576126; PMCID: PMC10418514.

[14] Hadden KB, Prince LY, Moore TD, James LP, Holland JR, Trudeau CR. Improving readability of informed consents for research at an academic medical institution. J Clin Transl Sci. 2017 Dec;1(6):361-365. doi: 10.1017/cts.2017.312. PMID: 29707258; PMCID: PMC5915809.

[15] O'Sullivan L, Sukumar P, Crowley R, McAuliffe E, Doran P. Readability and understandability of clinical research patient information leaflets and consent forms in Ireland and the UK: a retrospective quantitative analysis. BMJ Open. 2020 Sep 3;10(9):e037994. doi: 10.1136/bmjopen-2020-037994. PMID: 32883734; PMCID: PMC7473620.

[16] Joffe S, Cook EF, Cleary PD, Clark JW, Weeks JC. Quality of informed consent: a new measure of understanding among research subjects. J Natl Cancer Inst. 2001 Jan 17;93(2):139-47. doi: 10.1093/jnci/93.2.139. PMID: 11208884.

[17]: N. L. of Medicine. ClinicalTrials.gov API. https://clinicaltrials.gov/data-api/api. [Accessed 23-11-2023]. 2023

[18]: Artifex. PyMuPDF 1.24.9 documentation — pymupdf.readthedocs.io. https://pymupdf.readthedocs.io/en/latest/. [Accessed 22-11-2023]. 2023.

[19] Lecompte, L. L., and S. J. Young. 'Revised Common Rule Changes to the Consent Process and Consent Form. " En'. Ochsner J, vol. 20, 2020, pp. 62–75.

[20] ClinicalTrials.gov Identifier: NCT03923790. (n.d.). *A Study to Evaluate the Safety and Efficacy of [Study Intervention/Drug Name, if applicable]*. Retrieved from https://clinicaltrials.gov/study/NCT03923790?id=NCT03923790&rank=1

[21] Rahimli Ocakoglu S, Coskun B. The Emerging Role of AI in Patient Education: A Comparative Analysis of LLM Accuracy for Pelvic Organ Prolapse. Med Princ Pract. 2024 Mar 25;33(4):330–7. doi: 10.1159/000538538. Epub ahead of print. PMID: 38527444; PMCID: PMC11324208.

[22] Lambert, Raphaella, et al. "Assessing the Application of Large Language Models in Generating Dermatologic Patient Education Materials According to Reading Level: Qualitative Study." *JMIR dermatology* 7 (2024): e55898.

[23] Van Veen D, Van Uden C, Blankemeier L, Delbrouck JB, Aali A, Bluethgen C, Pareek A, Polacin M, Reis EP, Seehofnerová A, Rohatgi N, Hosamani P, Collins W, Ahuja N, Langlotz CP, Hom J, Gatidis S, Pauly J, Chaudhari AS. Clinical Text Summarization: Adapting Large Language Models Can Outperform Human Experts. Res Sq [Preprint]. 2023 Oct 30:rs.3.rs-3483777. doi: 10.21203/rs.3.rs-3483777/v1. Update in: Nat Med. 2024 Apr;30(4):1134-1142. doi: 10.1038/s41591-024-02855-5. PMID: 37961377; PMCID: PMC10635391.

[24] Tariq A, Urooj A, Trivedi S, et al. Patient Centric Summarization of Radiology Findings using Large Language Models[J]. medRxiv, 2024: 2024.02. 01.24302145.

[25] Lyu M, Peng C, Li X, et al. Automatic Summarization of Doctor-Patient Encounter Dialogues Using Large Language Model through Prompt Tuning[J]. arXiv preprint arXiv:2403.13089, 2024.

[26] Mathur Y, Rangreji S, Kapoor R, et al. Summqa at mediqa-chat 2023: In-context learning with gpt-4 for medical summarization[J]. arXiv preprint arXiv:2306.17384, 2023.

[27] Akash Ghosh, Mohit Tomar, Abhisek Tiwari, Sriparna Saha, Jatin Salve, and Setu Sinha. 2024. From Sights to Insights: Towards Summarization of Multimodal Clinical Documents. In *Proceedings of the 62nd Annual Meeting of the Association for Computational Linguistics (Volume 1: Long Papers)*, pages 13117–13129, Bangkok, Thailand. Association for Computational Linguistics.

[28] Liu Y, Ju S, Wang J. Exploring the potential of ChatGPT in medical dialogue summarization: a study on consistency with human preferences[J]. BMC Medical Informatics and Decision Making, 2024, 24(1): 75.

[29] White R, Peng T, Sripitak P, et al. CliniDigest: a case study in large language model based large-scale summarization of clinical trial descriptions[C]//Proceedings of the 2023 ACM Conference on Information Technology for Social Good. 2023: 396-402.

[30] Phatak A, Savage DW, Ohle R, Smith J, Mago V. Medical text simplification using reinforcement learning (TESLEA): Deep learning–based text simplification approach. JMIR Med Inform. JMIR Publications Inc.; 2022 Nov 18;10(11):e38095. PMID: 36399375

[31] Jeblick K, Schachtner B, Dexl J, Mittermeier A, Stüber AT, Topalis J, Weber T, Wesp P, Sabel B, Ricke J, Ingrisch M. ChatGPT makes medicine easy to swallow: An exploratory case study on simplified radiology reports [Internet]. arXiv [cs.CL]. 2022 [cited 2024 Jan 3]. Available from: http://arxiv.org/abs/2212.14882

[32] Goldsack T, Luo Z, Xie Q, Scarton C, Shardlow M, Ananiadou S, Lin C. BioLaySumm 2023 Shared Task: Lay Summarisation of Biomedical Research Articles. The 22nd Workshop on Biomedical Natural Language Processing and BioNLP Shared Tasks. Toronto, Canada: Association for Computational Linguistics; 2023. p. 468–477.

[33] Van Veen D, Van Uden C, Blankemeier L, Delbrouck JB, Aali A, Bluethgen C, Pareek A, Polacin M, Reis EP, Seehofnerová A, Rohatgi N, Hosamani P, Collins W, Ahuja N, Langlotz CP, Hom J, Gatidis S, Pauly J, Chaudhari AS. Clinical Text Summarization: Adapting Large Language Models Can Outperform Human Experts. Res Sq [Preprint]. 2023 Oct 30:rs.3.rs-3483777. doi: 10.21203/rs.3.rs-3483777/v1. Update in: Nat Med. 2024 Apr;30(4):1134-1142. doi: 10.1038/s41591-024-02855-5. PMID: 37961377; PMCID: PMC10635391.

[34] Decker H, Trang K, Ramirez J, et al. Large Language Model−Based Chatbot vs Surgeon-Generated Informed Consent Documentation for Common Procedures. *JAMA Netw Open.* 2023;6(10):e2336997. doi:10.1001/jamanetworkopen.2023.36997

[35] Ali, R., Connolly, I.D., Tang, O.Y. *et al.* Bridging the literacy gap for surgical consents: an AI-human expert collaborative approach. *npj Digit. Med.* 7, 63 (2024). https://doi.org/10.1038/s41746-024-01039-2

[36] Vaira LA, Lechien JR, Maniaci A, Tanda G, Abbate V, Allevi F, Arena A, Beltramini GA, Bergonzani M, Bolzoni AR, Crimi S, Frosolini A, Gabriele G, Maglitto F, Mayo-Yáñez M, Orrù L, Petrocelli M, Pucci R, Saibene AM, Troise S, Tel A, Vellone V, Chiesa-Estomba CM, Boscolo-Rizzo P, Salzano G, De Riu G. Evaluating AI-Generated informed consent documents in oral surgery: A comparative study of ChatGPT-4, Bard gemini advanced, and human-written consents. J Craniomaxillofac Surg. 2024 Oct 25:S1010-5182(24)00283-X. doi: 10.1016/j.jcms.2024.10.002. Epub ahead of print. PMID: 39490345.

[37] Mirza F N, Tang O Y, Connolly I D, et al. Using ChatGPT to facilitate truly informed medical consent[J]. NEJM AI, 2024, 1(2): AIcs2300145.

[38] Hereu P, Pérez E, Fuentes I, Vidal X, Suñé P, Arnau JM. Consent in clinical trials: what do patients know? Contemp Clin Trials. 2010 Sep;31(5):443–446. PMID: 20462521

[39] Juan-Salvadores P, Michel Gómez MS, Jiménez Díaz VA, Martínez Reglero C, Iñiguez Romo A. Patients' knowledge about their involvement in clinical trials. A non-randomized controlled trial. Front Med. 2022 Sep 20;9:993086. PMCID: PMC9531127

[40] Pietrzykowski T, Smilowska K. The reality of informed consent: empirical studies on patient comprehension-systematic review. Trials. 2021 Jan 14;22(1):57. PMCID: PMC7807905

[41] Peterson JS, Plana D, Bitterman DS, Johnson SB, Aerts HJWL, Kann BH. Growth in eligibility criteria content and failure to accrue among National Cancer Institute (NCI)-affiliated clinical trials. Cancer Med. 2023 Feb;12(4):4715–4724. PMCID: PMC99720315

[42] Unger JM, Vaidya R, Hershman DL, Minasian LM, Fleury ME. Systematic Review and Meta-Analysis of the Magnitude of Structural, Clinical, and Physician and Patient Barriers to Cancer Clinical Trial Participation. J Natl Cancer Inst. 2019 Mar 1;111(3):245–255. PMCID: PMC6410951

[43] Rimel BJ. Clinical Trial Accrual: Obstacles and Opportunities. Front Oncol. 2016 Apr 25;6:103. PMCID: PMC4843106

[44] Malmqvist E, Juth N, Lynöe N, Helgesson G. Early stopping of clinical trials: charting the ethical terrain. Kennedy Inst Ethics J. Johns Hopkins University Press; 2011 Mar;21(1):51–78. PMID: 21598846

[45] Schwartz AL, Alsan M, Morris AA, Halpern SD. Why Diverse Clinical Trial Participation Matters. N Engl J Med. 2023 Apr 6;388(14):1252–1254. PMID: 37017480

[46] Clark LT, Watkins L, Piña IL, Elmer M, Akinboboye O, Gorham M, Jamerson B, McCullough C, Pierre C, Polis AB, Puckrein G, Regnante JM. Increasing Diversity in Clinical Trials: Overcoming Critical Barriers. Curr Probl Cardiol. 2019 May;44(5):148–172. PMID: 30545650

[47] National Academies of Sciences, Engineering, and Medicine, Health and Medicine Division, Board on Population Health and Public Health Practice, Committee on Community-Based Solutions to Promote Health Equity in the United States. Communities in Action: Pathways to Health Equity. National Academies Press; 2017.

[48] Farb A, Viviano CJ, Tarver ME. Diversity in Clinical Trial Enrollment and Reporting-Where We Are and the Road Ahead. JAMA cardiology. 2023. p. 803–805. PMID: 37494022

[49] Murthy VH, Krumholz HM, Gross CP. Participation in cancer clinical trials: race-, sex-, and age-based disparities. JAMA. 2004 Jun 9;291(22):2720–2726. PMID: 15187053

[50] Mazzochi AT, Dennis M, Chun HYY. Electronic informed consent: effects on enrolment, practical and economic benefits, challenges, and drawbacks—a systematic review of studies within randomized controlled trials. Trials [Internet]. Springer Science and Business Media LLC; 2023 Feb 21 [cited 2024 Feb 5];24(1). Available from: https://pubmed.ncbi.nlm.nih.gov/36810093/ PMID: 36810093

[51] Golembiewski EH, Mainous AG III, Rahmanian KP, Brumback B, Rooks BJ, Krieger JL, Goodman KW, Moseley RE, Harle CA. An electronic tool to support patient-centered broad consent: A multi-arm randomized clinical trial in family medicine. Ann Fam Med. Annals of Family Medicine; 2021 Jan;19(1):16–23. PMID: 33431386

[52] De Sutter E, Borry P, Geerts D, Huys I. Personalized and long-term electronic informed consent in clinical research: stakeholder views. BMC Med Ethics [Internet]. Springer Science and Business

Media LLC; 2021 Dec 31 [cited 2024 Feb 5];22(1). Available from: https://pubmed.ncbi.nlm.nih.gov/34332572/ PMID: 34332572

[53] Lajonchere C, Naeim A, Dry S, Wenger N, Elashoff D, Vangala S, Petruse A, Ariannejad M, Magyar C, Johansen L, Werre G, Kroloff M, Geschwind D. An integrated, scalable, electronic video consent process to power precision health research: Large, population-based, cohort implementation and scalability study. J Med Internet Res. JMIR Publications Inc.; 2021 Dec 8;23(12):e31121. PMID: 34889741

[54] Synnot A, Ryan R, Prictor M, Fetherstonhaugh D, Parker B. Audio-visual presentation of information for informed consent for participation in clinical trials. Cochrane Libr [Internet]. Wiley; 2014 May 9 [cited 2024 Feb 5];2014(5). Available from: https://pubmed.ncbi.nlm.nih.gov/24809816/ PMID: 24809816

[55] Kim Y J, DeLisa J A, Chung Y C, et al. Recruitment in a research study via chatbot versus telephone outreach: a randomized trial at a minority-serving institution[J]. Journal of the American Medical Informatics Association, 2022, 29(1): 149-154.

[56] Savage SK, LoTempio J, Smith ED, Andrew EH, Mas G, Kahn-Kirby AH, Délot E, Cohen AJ, Pitsava G, Nussbaum R, Fusaro VA, Berger S, Vilain E. Using a chat-based informed consent tool in large-scale genomic research. J Am Med Inform Assoc. Oxford University Press (OUP); 2024 Jan 18;31(2):472–478. PMID: 37665746

[57] Goodman KE, Yi PH, Morgan DJ. AI-Generated Clinical Summaries Require More Than Accuracy. *JAMA*. 2024;331(8):637–638. doi:10.1001/jama.2024.0555

[58] Bitterman DS, Aerts HJWL, Mak RH. Approaching autonomy in medical artificial intelligence. Lancet Digit Health. 2020 Sep;2(9):e447-e449. doi: 10.1016/S2589-7500(20)30187-4. PMID: 33328110.

[59] Kim J, Chen ML, Rezaei SJ, et al. Perspectives on Artificial Intelligence–Generated Responses to Patient Messages. *JAMA Netw Open.* 2024;7(10):e2438535. doi:10.1001/jamanetworkopen.2024.38535

[60] Mannhardt N, Bondi-Kelly E, Lam B, et al. Impact of Large Language Model Assistance on Patients Reading Clinical Notes: A Mixed-Methods Study[J]. arXiv preprint arXiv:2401.09637, 2024.

[61] Tam, T.Y.C., Sivarajkumar, S., Kapoor, S. *et al.* A framework for human evaluation of large language models in healthcare derived from literature review. *npj Digit. Med.* 7, 258 (2024). https://doi.org/10.1038/s41746-024-01258-7

# Appendix

Table 1. 11 clinical trial ICFs from ClinialTrials.gov

| S. No. | NCT number | Clinical Trial name |
|---|---|---|
| 1 | NCT03041090 | Patterns and Prevalence of FDG Extravasation in PET/CT Scans (Lucerno device) |
| 2 | NCT03572790 | Effects of Seven Day Prucalopride Administration in Healthy Volunteers |
| 3 | NCT03871556 | Diagnostic Ultrasound for Measuring Fat of the Body |
| 4 | NCT03923790 | Stroke Telemedicine Outpatient Prevention Program for Blood Pressure Reduction (STOP-Stroke) |
| 5 | NCT04001790 | Effect of a 9-Month Internship Intervention for Military Dependents With ASD |
| 6 | NCT04044456 | Combining Attention and Metacognitive Training to Improve Goal Directed Behavior in Veterans With TBI |
| 7 | NCT04122456 | DNA Repair Activity in the Skin of Day and Night Shift Workers |
| 8 | NCT04152603 | Better Research Interactions for Every Family (BRIEF) |
| 9 | NCT05253703 | Immersive Virtual Reality Bicycling for Healthy Adults |
| 10 | NCT05312190 | Clinical and Basic Researches Related to ZhenQi Buxue Oral Liquid in Treating Menstrual Disorders |
| 11 | NCT05713903 | Laparoscopic Versus Open Right Colectomy for Right Colon Cancer |

Table 2. Informed consent elements extracted for sequential prompting approach

| Topic* | Text used in prompt to extract relevant text from ICF |
|---|---|
| A statement that the study involves research | Does the study involve medical research? |
| An explanation of the purposes of the research | What is the purpose of this research study? |
| The expected duration of the subject's participation | How long will the participant be involved in the study? |
| A description of the procedures to be followed | What procedures will the participant need to follow? |
| Identification of any procedures which are experimental | Are any of the procedures experimental? |
| A description of any reasonably foreseeable risks or discomforts to the subject | What are the risks or discomforts of participating? |
| A description of any benefits to the subject or to others which may reasonably be expected from the research | What are the benefits of participating? |
| The approximate number of subjects involved in the study | How many people will be participating in the study? |
| A disclosure of appropriate alternative procedures or courses of treatment, if any, that might be advantageous to the subject | What other treatment options may help the participant besides this research study? |
| A statement describing the extent, if any, to which confidentiality of records identifying the subject will be maintained | How will the researchers keep the participant's records private? |
| For research involving more than minimal risk, an explanation as to whether any compensation, and an explanation as to whether any medical treatments are available, if injury occurs and, if so, what they consist of, or where further information may be obtained | What are details on compensation, medical treatments, injuries, and where individuals can find more information? |
| Research, Rights or Injury: An explanation of whom to contact for answers to pertinent questions about the research and research subjects' rights, and whom to contact in the event of a research-related injury to the subject | Who can the participant contact if they have questions or get hurt in the research? |
| A statement that participation is voluntary, refusal to participate will involve no penalty or loss of benefits to which the subject is otherwise entitled, and the subject may discontinue participation at any time without penalty or loss of benefits, to which the subject is otherwise entitled | Can the participant drop out of the study anytime without consequences? |
| Anticipated circumstances under which the subject's participation may be terminated by the investigator without regard to the subject's or the legally authorized representative's consent | For what reasons could the researchers remove the participant from the study? |
| Any additional costs to the subject that may result from participation in the research | Will participating cost the participant anything? |
| The consequences of a subject's decision to withdraw from the research and procedures for orderly termination of participation by the subject | If the participant wants to stop the study, what should they do and what happens next? |
| A statement that significant new findings developed during the course of the research that may relate to the subject's willingness to continue participation will be provided to the subject | Will the researchers let the participant know if they find anything important that could change their decision to stay in the study? |

*Topics are based on basic elements of informed consent in Subpart A of the Revised Common Rule
ICF = informed consent form

Table 3. 5 clinical trial ICFs for patient survey evaluation

| Trial No. | NCT number | Clinical Trial name |
|---|---|---|
| 1 | N/A (Internal, non-therapeutic departmental trial) | BWH Radiation Oncology All-Department Biorepository to Accelerate New Discoveries (BROADBAND), in collaboration with the Mass General Brigham Biobank |
| 2 | NCT04267848 | Testing the Addition of a Type of Drug Called Immunotherapy to the Usual Chemotherapy Treatment for Non-Small Cell Lung Cancer, ALCHEMIST Trial |
| 3 | NCT04037462 | Induction of Senescence using Dexamethasone to Re-sensitize NSCLC to Anti-PD1 Therapy |
| 4 | NCT04091165 | Mobile APP Utilization for Enhanced Post-Operative Nutritional Recovery |
| 5 | NCT04542291 | Targeting Pancreatic Cancer With Sodium Glucose Transporter 2 (SGLT2) Inhibition |

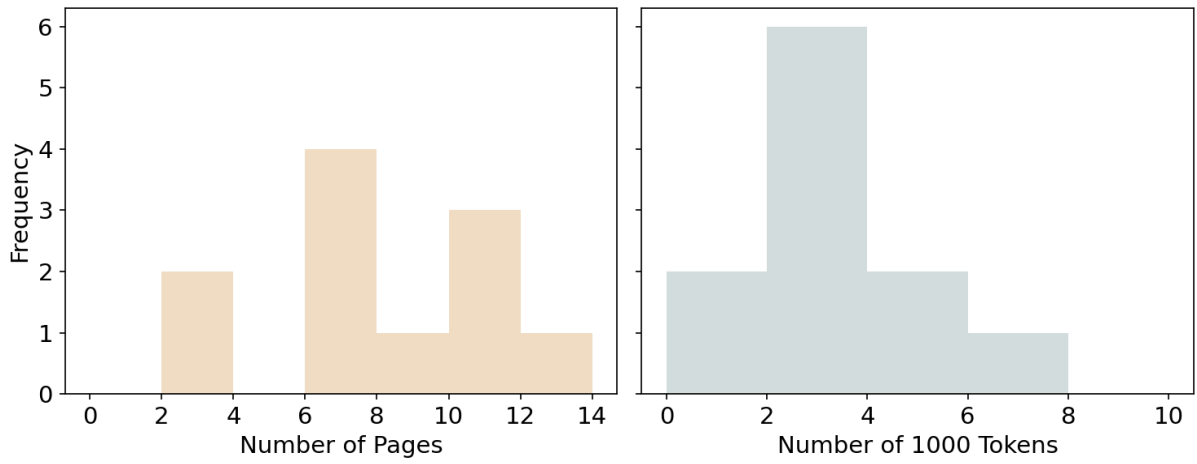

Figure 1. Length distribution of 11 ICFs for summary generation: pages (left) and tokens (right).

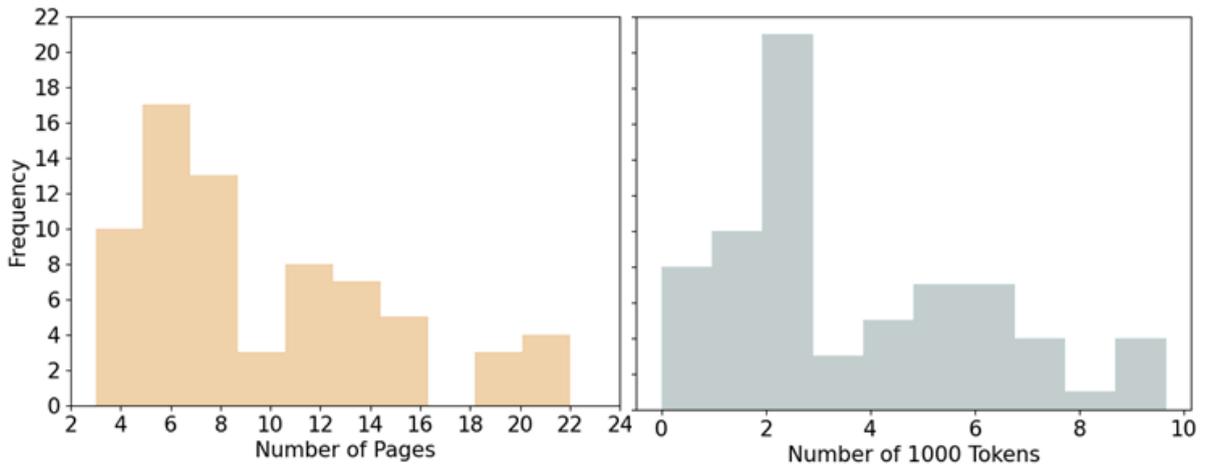

Figure 2. Length distribution of 91 ICFs for MCQAs generation: pages (left) and tokens (right)

> As an intelligent principal investigator of a clinical trial, you must provide a clear summary using the consent form text. The summary should include a statement that explain the purpose of the research, a description of the procedures, how long the subject will be involved in the study, the risks, the benefits, the subject's participation is voluntary, and the alternatives if the subject decides not to participate.
>
> Use the following guidelines to create a summary in a paragraph style:
>
> - Summary must be at most **150 words**
> - Simplify any complex terms or concepts
> - Make the summary highly understandable, recommended for **eighth-grade** level audience
> - Use respectful and empowering language for patients
> - Spell out acronyms upon first use
> - Include all relevant information without adding extra details
> - Use active voice
> - Keep the summary concise and to the point
>
> Form Text: **{form_text}**

Figure 3. Prompt used for direct summarization from informed consent form text

(A)

You are a smart principal investigator for a clinical trial, and I will give you the text of the consent form. Below are questions separated by |. I want you to get the full exact text for each question if it's in the form else return "na". Provide responses with the exact text from the form in defined format. Rephrasing text is not allowed, and you should not skip over information on the form.

Points:
[1. Does the study involve medical research?|
2. What is the purpose of this research study?|
3. How long will the participant be involved in the study?|
4. What procedures will the participant need to follow?|
5. Are any of the procedures experimental?|
6. What are the risks or discomforts of participating?|
7. What are the benefits of participating?|
8. How many people will be participating in the study?]

response1 = {{"study_research": **<exact_text_1>**,
    "purpose": **<exact_text_2>**,
    "duration": **<exact_text_3>**,
    "procedures": **<exact_text_4>**,
    "experimental_procedures": **<exact_text_5>**,
    "risks": **<exact_text_6>**,
    "benefits": **<exact_text_7>**,
    "participants": **<exact_text_8>**}}

Be consistent with the naming of the keys and the format of the values. Form text: **{form_text}**

---

You are a smart principal investigator for a clinical trial, and you will be given the text of consent form. Below are questions separated by |. I want you to get the full exact text for each question if it's in the form else return "na". Provide responses with the exact text from the form in the defined format. Rephrasing text is not allowed, and you should not skip over information on the form.

Points:
[ 9. What other treatment options may help the participant besides this research study?|
 10. How will the researchers keep the participant's records private?|
 11. What are details on compensation, medical treatments, injuries, and where individuals can find more information?|
 12. Who can the participant contact if they have questions or get hurt in the research?|
 13. Can the participant drop out of the study anytime without consequences?|
 14. For what reasons could the researchers remove the participant from the study?|
 15. Will participating cost the participant anything?|
 16. If the participant wants to stop the study, what should they do and what happens next?|
 17. Will the researchers let the participant know if they find anything important that could change their decision to stay in the study?]

response2 = {{"alternative_procedures": **<exact_text_1>**,
    "confidentiality": **<exact_text_2>**,
    "compensation": **<exact_text_3>**,
    "contact_info": **<exact_text_4>**,
    "voluntary_participation": **<exact_text_5>**,
    "discontinue_cooperation": **<exact_text_6>**,
    "additional_costs": **<exact_text_7>**,
    "withdrawal_effects": **<exact_text_8>**,
    "new_findings": **<exact_text_9>**}}

Be consistent with the naming of the keys and the format of the values. Form text: **{form_text}**

(B)

> As the intelligent principal investigator of a clinical trial, you've received extracted text from the consent form. You must provide a clear summary.
>
> The summary should include a statement that explain the purpose of the research, a description of the procedures, how long the subject will be involved in the study, the risks, the benefits, the subject's participation is voluntary, and the alternatives if the subject decides not to participate.
>
> Use the following guidelines to create a summary in a paragraph style:
>
> - Summary must be at most **150 words**
> - Simplify any complex terms or concepts
> - Make the summary highly understandable, recommended for **eighth-grade** level audience
> - Use respectful and empowering language for patients
> - Spell out acronyms upon first use
> - Include all relevant information without adding extra details
> - Use active voice
> - Keep the summary concise and to the point
>
> Your extracted text is: **{extracted_content}**

Figure 4. Prompts used for sequential summarization from informed consent form text. A) Prompts used to extract text sections describing elements of informed consent. B) Prompt used to generate a summary based on the extracted text sections.

> **System Prompt**: 'You are a smart AI assistant. Given an example consent form and a multiple-choices question in a specific topic regarding the form that helps patient understand the form. Now generate new questions regarding the new form in the same topic.'
>
> **User Prompt 1**: '===Example consent form===: \n <text of the example ICF> \n\n ===Topic===: \n <target topic> \n Generate a multiple-choices question in the given topic regarding the given consent form.'
>
> **Assistant Prompt**: '===Example question===: \n <human-generated MCQA for the target topic>'
>
> **User Prompt 2**: '===New consent form===: \n <target ICF> \n\n ===Topic===: \n <target topic> \n Generate one multiple-choices question in the given topic regarding the new consent form; the correct option of each new question should not be the original sentences from the consent form.'
>
> ```
> message_text = [
>     {"role": "system", "content": <System Prompt> },
>     {"role": "user", "content": <User Prompt 1> },
>     {"role": "assistant", "content": <Assistant Prompt> },
>     {"role": "user", "content": <User Prompt 2> }
> ]
> ```

Figure 5. Multi-turn prompts for generating MCQAs based on in-context learning

```
System Prompt: You are an AI assistant for answering the multiple-choices question

User Prompt: Please answer the following multiple-choice question based on the given
clinical trial consent form; there is only one correct answer. Return the correct option at
the beginning of the response, then explain reason and why other options are
incorrect. \n\n\n  Consent form: <target ICF> \n\n\n Question: <MCQA>
message_text = [
      {"role": "system", "content": <System Prompt> },
      {"role": "user", "content": <User Prompt> },
]
```

Figure 6. Prompt used to cross-reference the GPT4-generated MCQAs with 4 API LLMs (GPT-4o, Cohere R+, Gemini Pro 1 and Claude 3 Sonnet).

Table 4. Self-reported Information of 504 Annotators for MCQAs evaluation

| Experience Level | Count |
|---|---|
| MD | 13 |
| DO | 1 |
| NP | 7 |
| NP Student | 10 |
| Other | 78 |
| Other Healthcare Students | 40 |
| PA | 4 |
| PA Student | 4 |
| Pharmacist | 2 |
| Pharmacy Student | 4 |
| Not reported | 261 |

Table 5. Complete Results of Patient Surveys of Clinical Trial Summaries

| Survey Item | Strongly disagree | Disagree | Neither agree or disagree | Agree | Strongly agree | Missing | | | |
|---|---|---|---|---|---|---|---|---|---|
| **BROADBAND Summary** | | | | | | | | | |
| This summary is easy to understand. | 0 | 0 | 2 | 4 | 6 | 1 | | | |
| If I were researching trials, this summary provides enough information for me to decide if I would want to contact the research team to learn more about the trial. | 0 | 0 | 4 | 5 | 4 | 0 | | | |
| I believe that reading this summary improved my understanding of BROADBAND | 0 | 0 | 2 | 6 | 5 | 0 | | | |
| Were there topics you found hard to understand during the BROADBAND consent discussion? | Purpose of the study | Study procedures | Duration of participation | Risks of particpating | Benefits of participating | Whether the study is voluntary | Alternatives to participating | Other | |
| | 0 | 0 | 0 | 0 | 0 | 0 | 0 | 1 | |
| **Trial 2 Summary** | | | | | | | | | |
| Survey Item | Strongly disagree | Disagree | Neither agree or disagree | Agree | Strongly agree | Missing | | | |
| This summary is easy to understand. | 0 | 2 | 0 | 3 | 5 | 3 | | | |
| If I were researching trials, this summary provides enough information for me to decide if I would want to contact the research team to learn more about the trial. | 0 | 1 | 1 | 5 | 4 | 2 | | | |
| Which version of the trial do you prefer? | Version 1 (paragraph) | Version 2 (list) | No preference | Missing | | | | | |
| | 1 | 8 | 1 | 3 | | | | | |
| **Trial 3 Summary** | | | | | | | | | |
| Survey Item | Strongly disagree | Disagree | Neither agree or disagree | Agree | Strongly agree | Missing | | | |
| This summary is easy to understand. | 0 | 1 | 0 | 4 | 5 | 3 | | | |
| If I were researching trials, this summary provides enough information for me to decide if I would want to contact the research team to learn more about the trial. | 0 | 1 | 1 | 4 | 5 | 2 | | | |
| **Trial 4 Summary** | | | | | | | | | |
| Survey Item | Strongly disagree | Disagree | Neither agree or disagree | Agree | Strongly agree | Missing | | | |
| This summary is easy to understand. | 0 | 0 | 1 | 3 | 5 | 4 | | | |
| If I were researching trials, this summary provides enough information for me to decide if I would want to contact the research team to learn more about the trial. | 0 | 0 | 2 | 4 | 4 | 3 | | | |
| **Trial 5 Summary** | | | | | | | | | |
| Survey Item | Strongly disagree | Disagree | Neither agree or disagree | Agree | Strongly agree | Missing | | | |
| This summary is easy to understand. | 0 | 0 | 0 | 4 | 5 | 5 | | | |
| If I were researching trials, this summary provides enough information for me to decide if I would want to contact the research team to learn more about the trial. | 0 | 0 | 2 | 3 | 4 | 4 | | | |

Table 6. Statistics of human evaluation on GPT4-generated MCQAs.

|  | Qualified Reads | Difficulty | Agreement |
|---|---|---|---|
| **Mean** | 5.207 | 0.155 | 0.869 |
| **Std** | 1.988 | 0.221 | 0.172 |
| **Median** | 5.000 | 0.000 | 1.000 |
| **Min** | 1.000 | 0.000 | 0.250 |
| **Max** | 12.00 | 1.000 | 1.000 |
| **5% Quartile** | 2.000 | 0.000 | 0.500 |
| **10% Quartile** | 3.000 | 0.000 | 0.600 |
| **90% Quartile** | 8.000 | 0.500 | 1.000 |
| **95% Quartile** | 9.000 | 0.600 | 1.000 |

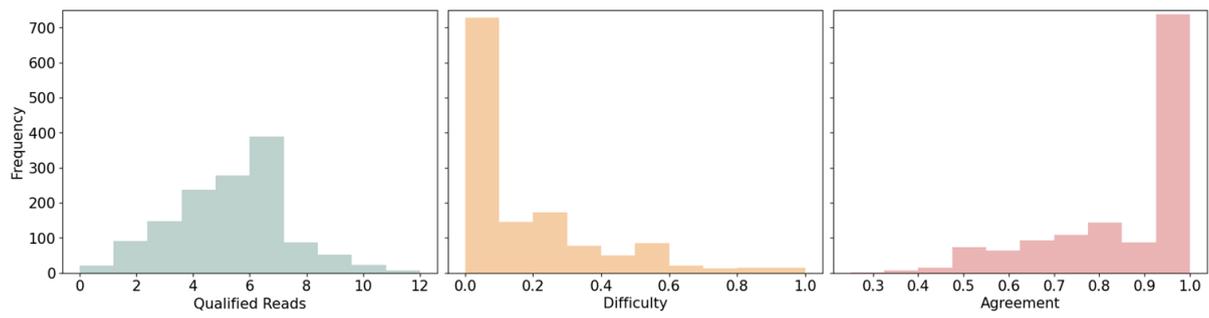

Figure 7. Distribution of Qualified Reads (left), Difficulty (middle) and Agreement (right) of 1335 MCQAs

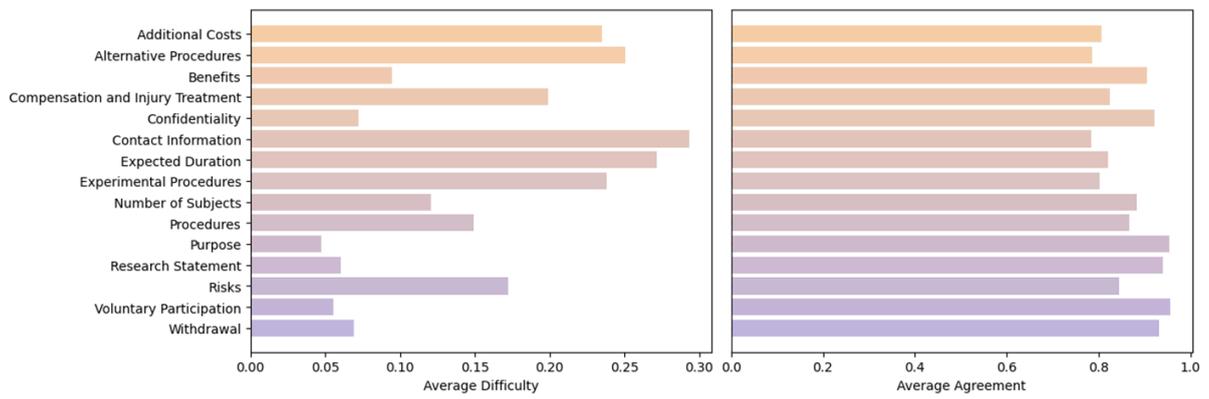

Figure 8. Average Difficulty (left) and Agreement (right) of MCQAs according to informed consent topics.

Table 6. The mapping between NIH Common Rules and short term in Appendix Figure 8

| Topic | Short Term |
|---|---|
| A statement that the study involves research | Research Statement |
| An explanation of the purposes of the research | Purpose |
| The expected duration of the subject's participation | Expected Duration |
| A description of the procedures to be followed | Procedures |
| The approximate number of subjects involved in the study | Number of Subjects |
| Identification of any procedures which are experimental | Experimental Procedures |
| A description of any reasonably foreseeable risks or discomforts to the subject | Risks |
| A description of any benefits to the subject or to others which may reasonably be expected from the research | Benefits |
| A disclosure of appropriate alternative procedures or courses of treatment, if any, that might be advantageous to the subject | Alternative Procedures |
| A statement describing the extent, if any, to which confidentiality of records identifying the subject will be maintained | Confidentiality |
| For research involving more than minimal risk, an explanation as to whether any compensation, and an explanation as to whether any medical treatments are available, if injury occurs and, if so, what they consist of, or where further information may be obtained | Compensation and Injury Treatment |
| Research, Rights or Injury: An explanation of whom to contact for answers to pertinent questions about the research and research subjects' rights, and whom to contact in the event of a research-related injury to the subject | Contact Information |
| A statement that participation is voluntary, refusal to participate will involve no penalty or loss of benefits to which the subject is otherwise entitled, and the subject may discontinue participation at any time without penalty or loss of benefits, to which the subject is otherwise entitled | Voluntary Participation |
| Any additional costs to the subject that may result from participation in the research | Additional Cost |
| The consequences of a subject's decision to withdraw from the research and procedures for orderly termination of participation by the subject | Withdraw |

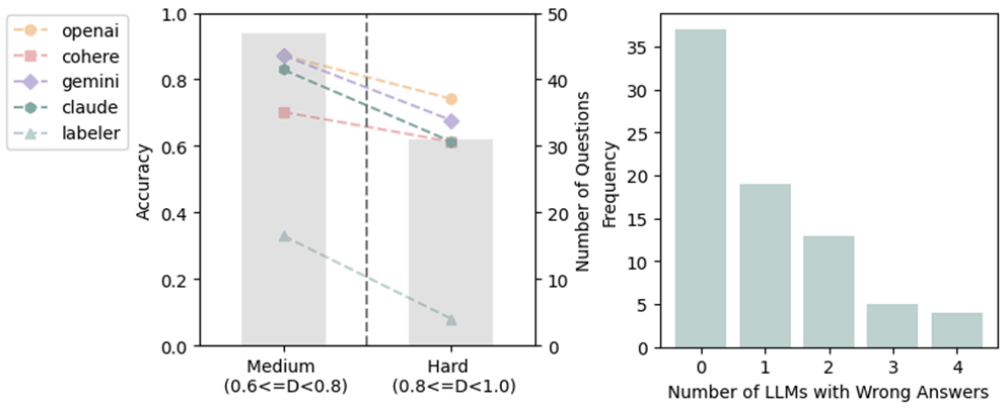

Figure 9. Distribution of Medium and Hard MCQAs and accuracy of four LLMs (GPT-4o, Cohere R+, Gemini Pro 1, Claude 3 Sonnet) and crowdsourced readers.

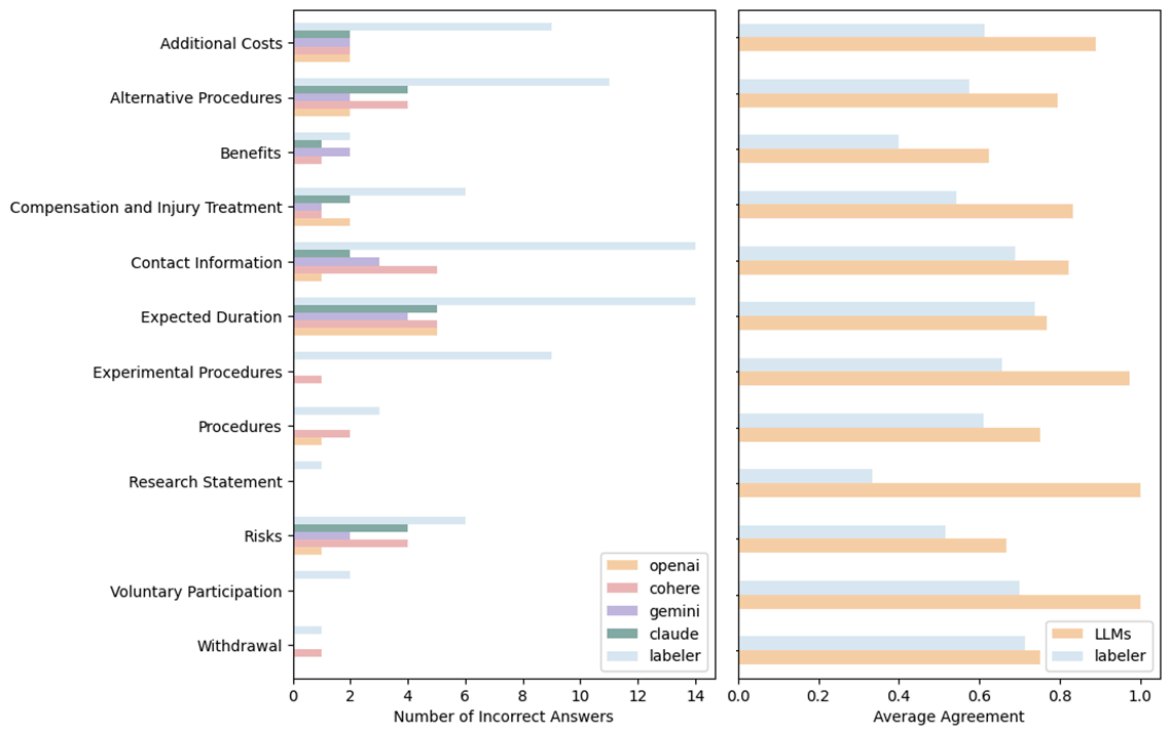

Figure 10. The number of MCQAs that each of LLM and readers give the incorrect answer (left) and the agreement of readers and 4 LLMs (right) on each NIH topic among the quality assurance set (difficulty ≥ 0.6 and agreement ≤ 0.5).

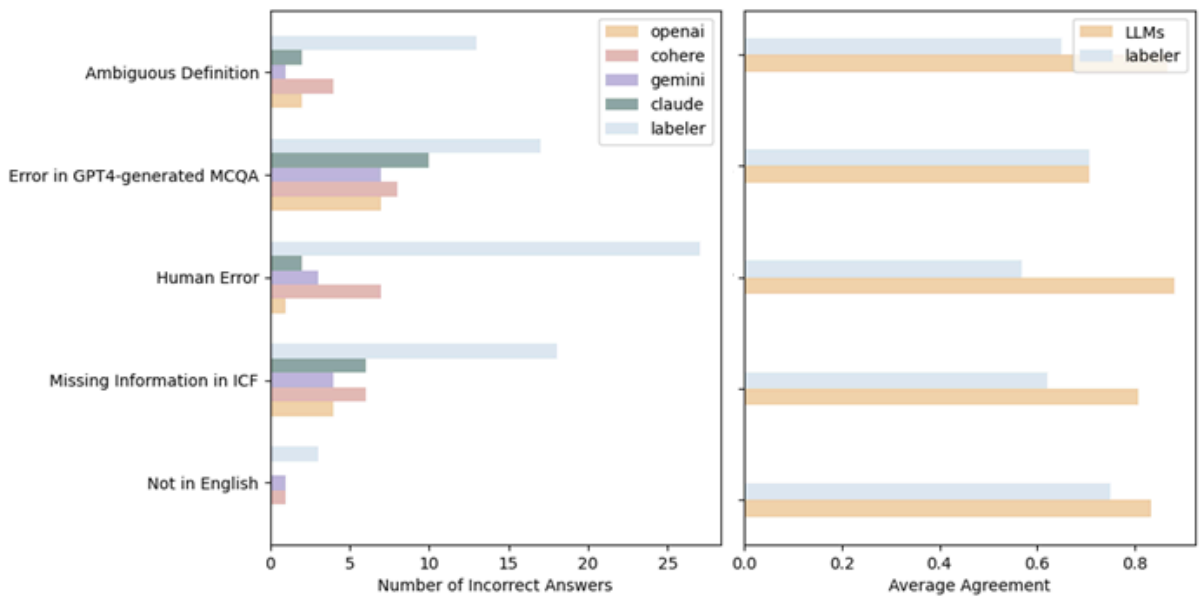

Figure 11. The number of MCQAs corresponding to each error mode (left) and the agreement of readers and 4 LLMs (right) on each error mode shown in Table 4 among the quality assurance set (difficulty ≥ 0.6 and agreement ≤ 0.5).

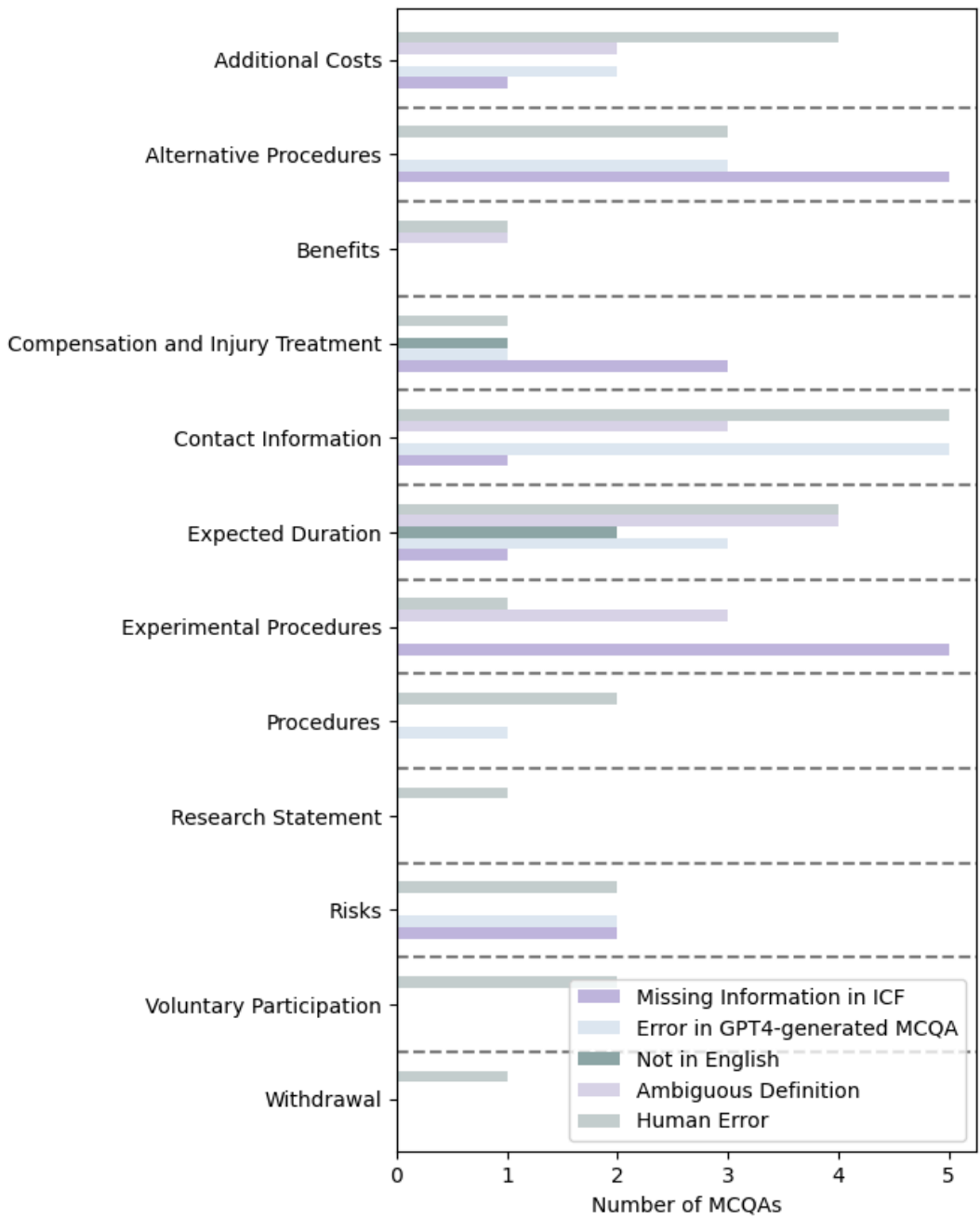

Figure 12. The number of MCQAs corresponding to each error mode that the readers gave the incorrect answer for, categorized according to topic in the quality assurance set (difficulty ≥ 0.6 and agreement ≤ 0.5).